\documentclass[letterpaper, 10 pt, conference]{ieeeconf}  

\IEEEoverridecommandlockouts                              

\usepackage{graphicx} 
\usepackage{mathptmx} 
\usepackage{times} 
\usepackage{amsmath} 
\usepackage{amssymb}  

\usepackage{caption}
\usepackage{subcaption}
\usepackage[english, status=draft]{fixme}
\fxusetheme{color}

\usepackage{makecell}

\title{\LARGE \bf
An End-to-End Robot Architecture to Manipulate\\ 
Non-Physical State Changes of Objects}

\author{Wonjun Yoon$^{*}$ \and Sol-A Kim$^{*}$ \and Jaesik Choi
\thanks{$^{*}$These authors contributed equally to this work.}
\thanks{School of Electrical and Computer Engineering, 
Ulsan National Institute of Science and Technology, Ulsan 44919, Republic of Korea
{\tt\small \{wjyoon26,johnksy,jaesik\}@unist.ac.kr}}
}

\begin{document}

\maketitle

\thispagestyle{plain}
\pagestyle{plain}

\begin{abstract}
With the advance in robotic hardware and intelligent software, humanoid robot plays an important role in various tasks including services for human assistance and difficult jobs for hazardous industries. Recent advances in task learning enable humanoid robots to conduct dexterous manipulation tasks such as grasping objects and assembling parts of furniture. Operating objects without physical movements is an even more challenging task for humanoid robot because (1) effects of actions may not be clearly seen in the physical configuration space and (2) meaningful actions could be very complex in a long time horizon. As an example, playing a mobile game on a smart device has such challenges because both swipe actions and complex state transitions inside the smart devices in a long time horizon. In this research paper, we solve this problem by introducing an integrated architecture which connects end-to-end dataflow from sensors to actuators in a humanoid robot to operate smart devices. We implement our integrated architecture in the Baxter Research Robot and experimentally demonstrate that the robot with our architecture could play a challenging mobile game called \textit{"2048"}, as accurate as in a simulated environment. 
\end{abstract}

\section{INTRODUCTION}
Manipulating objects with a humanoid robot is an important task since such human-like robots play an important role in service for human assistance and difficult jobs for hazardous industries in the near future. However, designing and building a proper robotic manipulation task is not trivial since the dynamics of the robot and constraints of objects should be carefully considered. Thus, a successful robotics manipulation task is not easily expanded to general human-like tasks such as assembling parts and smart device manipulation. Recent advances in a task learning under unknown dynamics enable humanoid robots to conduct various dexterous manipulation tasks such as assembling parts of furniture \cite{Control} and toys \cite{toy}. 

To achieve human-level manipulation, humanoid robots need to handle objects with no salient physical movement. As an example, touch-enabled smart devices including smart phones, tablet computers, and the Internet of Things (IoT) devices have widespread applications in everyday life. Compared to mechanical manipulations (e.g., device switches and buttons), touch-enabled smart devices are easier for developers to include rich functionality and for users to manipulate without much physical forces applied. Thus, a general-purpose humanoid robot may require to be able to learn (1) how to learn and evaluate actions in smart devices and (2) how to execute a long sequence of actions to fulfill a non-physical task goal. 

Manipulating touch-enabled smart devices has unique challenges in that such devices are usually fragile and require a long sequence of manipulation actions, i.e., a complex task planning. The manipulation becomes even harder especially under unknown dynamics of the robotics manipulator and unknown task constraints.  

We solve this issue by introducing a new integrated architecture which seamlessly connects dataflow from sensors to actuators, and satisfies the requirement of the local manipulations (swipes) and the long-term tasks.  We present a general framework to learn to fulfill complex tasks on smart devices. As shown in Fig. 2, our architecture includes general tools for recognition, planning, and execution. One important aspect of our architecture is that we could learn (or train) individual components to fulfill the goal of complex tasks. In detail, our architecture includes (1) learning task actions with long time horizons by deep reinforcement learning \cite{AtariPaper} and (2) learning manipulation actions from Linear Quadratic Regulator (LQR) \cite{LQRIROS15}. 

\begin{figure}[t]
\centering
\includegraphics[width=0.45\textwidth]{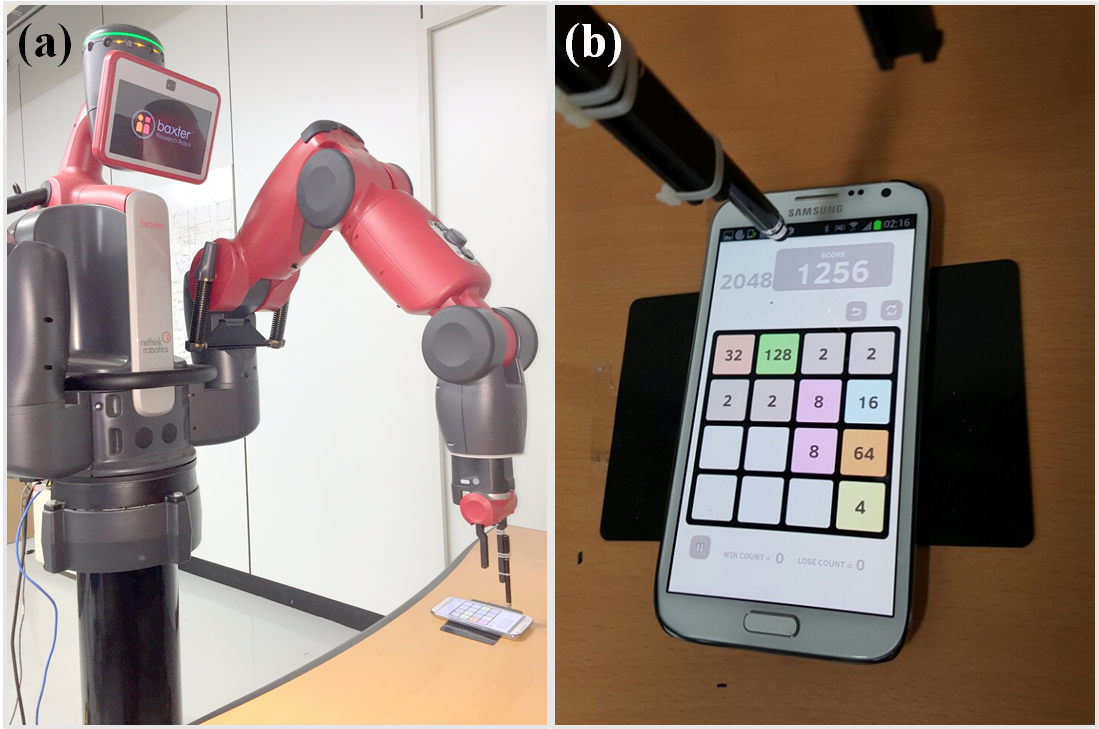}
\caption{(a): Our Baxter research robot. (b): The end effector and the smart device just before starting the \textit{"2048"} game.}
\label{fig:overview}
\end{figure}

We experimentally demonstrate that the integrated architecture can learn non-trivial manipulation tasks for smart devices. Specifically, we demonstrate that the Baxter Research robot \cite{baxter} equipped with our architecture can solve a non-trivial mobile game application called \textit{"2048"} \cite{the2048}, as shown in Fig. \ref{fig:overview}. In the experiments, we show that the winning rate of our Baxter robot for achieving the number 128 and 256 is as high as the Deep Q Learning \cite{AtariPaper} trained solely in a simulated environment. 

In the following, Section~\ref{sec:related_work} explains related work for learning manipulation tasks for humanoid robots. Section~\ref{sec:background} explains existing models used in our work; LQR and Deep Reinforcement Learning (DRL). Section~\ref{sec:learning_architecture} presents our architecture for the task learning. Section~\ref{sec:experiments} reports the experiments results followed by conclusion in Section~\ref{sec:conclusion}

\section{RELATED WORK}
\label{sec:related_work}

\begin{figure*}[th!]
\includegraphics[width=\textwidth]{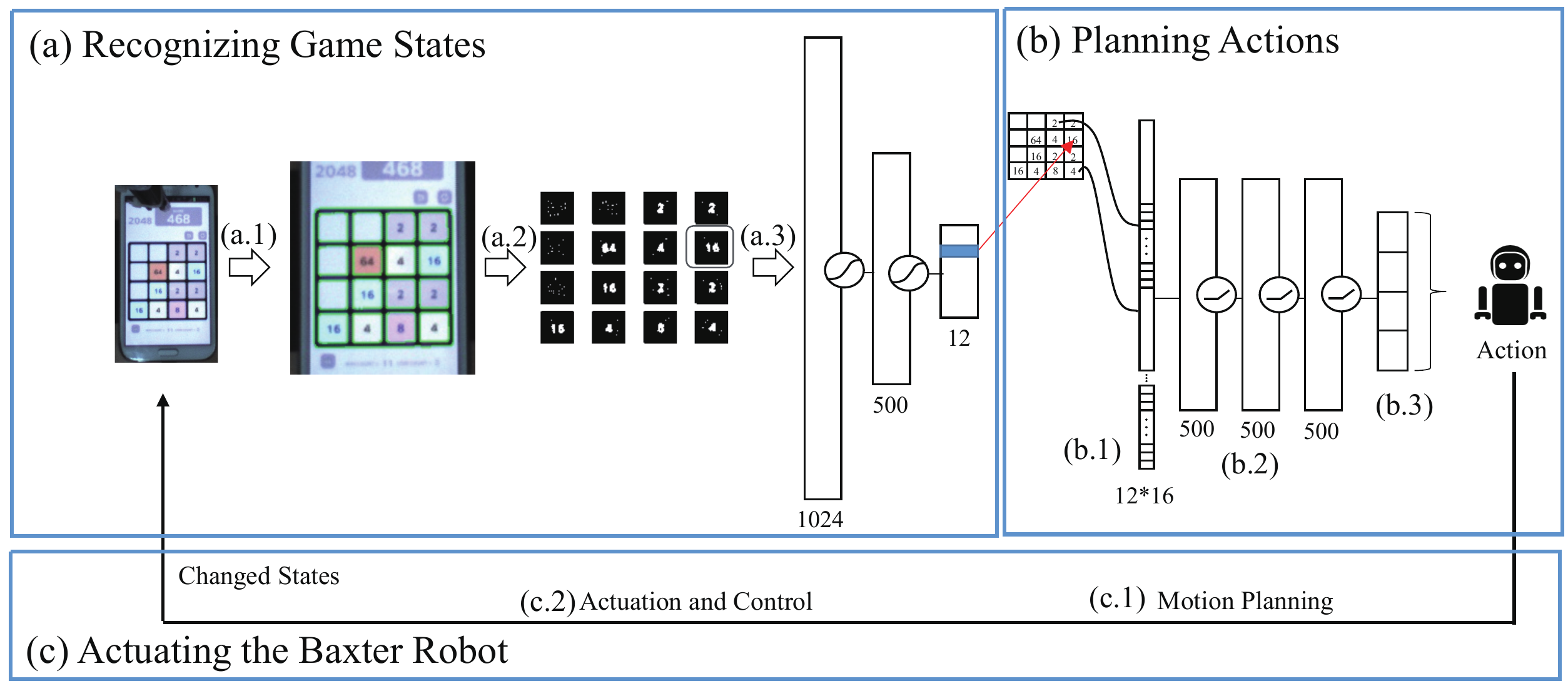}
\caption {An illustration of our architecture diagram. (a) presents the image recognition process: (a.1) detection of boundary by the Canny edge detector; (a.2) extraction of  tiles, 32x32 pixels, guided by the edge with the adaptive Gaussian filter \cite{adaptivegaussian}; and (a.3) recognition of numbers in each tile by a neural network with a hidden layer of 500 nodes where input is 1-dimensional input data and output is 12 binary values one for one of $\{\emptyset, 2, \cdots, 2048 \}$ 
 (b) presents learning action policy by Deep Q learning model: (b.1) a bit string mapped from the matrix, divided by 12 sections in ascending order from 0 to 2048 where each slots from each sections is filled with 0 or 1 to indicate the value of the tile in the specific positions among 16 spaces are in or not; (b.2) the neural network consists of three layers with fully connected layer, each has 500 hidden nodes, and the transfer function is the rectified unit, ReLu; and (b.3) four actions: Up, Down, Left, and Right. (c) presents the actuation of the robot: (c.1) optimization of four trajectories (left, right, backward, forward swipes) by iLQR controller, and (c.2) following the optimized trajectory with the direction guided by (c).}
\end{figure*}

Industrial robotic manipulators are used to verify the functionality of touch-enabled smart devices \cite{intel_robot,tapster,chameleon,optofidelity}.  
Commercial robotic manipulators are successfully used to test and verify the functionality of smart devices  
 \cite{chameleon,optofidelity}. Most of the commercial tests are designed to verify the functional characteristics of touches such as latency, sensed forces and durability with a specialized mobile application. Our work further extends these efforts to let a general-purpose humanoid robot manipulate non-physical state changes of smart devices to fulfill the requirement of complex tasks in long time horizons. 

LQR provides an optimal control solution where the system is a linear differential dynamic system and the cost function can be represented as a quadratic function. Under unknown dynamics, LQR is especially useful since LQR learns a locally linear action model which is less dependent on dynamics. Also, the linear Gaussian transition model is simple and easy to control. Thus, LQR has widely used to learn manipulation actions of robots \cite{LQR_ex1,LQR_ex2,LQR_ex3}. LQR does not require one to fully specify the dynamical constraints of the task and the robot manipulation. LQR learns a set of proper actions with minimal human (engineer) specifications. In experiments, we find a close-to-optimal actions by the iterative LQR \cite{iLQR,MPC_iLQR,LQRIROS15} which has been successful to learn various non-linear robotic manipulations.

Reinforcement learning has been extensively used in high-level tasks such as playing games \cite{AtariPaper}, planning to manage supply chains and controlling multi-agents in military systems. Reinforcement learning also holds the promise of enabling robots to learn motion skills for manipulating external objects. Recent advances have shown that it could learn various high-level tasks including playing tennis \cite{Tennis10}, stacking blocks \cite{Deisenroth11}, assembling parts using tools \cite{LQRIROS15}. 

There has been long efforts to learn low-level (motion planning) actions to conduct high-level (complex) tasks, which require to reason not only about the kinematic and geometric constraints but also intermingled constraints on state spaces \cite{KaelblingL13,KonidarisKL14,taskmotion,CPMP09}. Our ultimate goal is to automate the procedure of learning low-level actions and high-level tasks simultaneously. In this research paper, we present an integrate architecture to demonstrate the feasibility of our goal to learn motion planning and task planning, and then execute them together in an integrated humanoid platform.

\section{Background}
\label{sec:background}

\subsection{Linear Quadratic Regulator Models}
LQR solves feedback control problems where the system dynamics is composed of linear differential equations and the cost is represented as a quadratic function, called the LQR problem. The following equations are state and cost functions with a finite-horizon, discrete-time LQR,
\begin{align}
x_{t+1} &= A_tx_t + Bu_t, \label{eqn:lqr_trans}\\
l(x,u) &= \sum_{t=0}^N \left( x_t^TQx_t + u_t^TRu_t \right), 
\end{align}
where $x_t$ and $u_t$ are states and user inputs (control) respectively, and $Q$, $R$ and $N$ are predefined model parameters for the cost (or loss) function. Note that, we wish to find the optimal trajectory, $\tau = (x_0,u_0, \cdots, x_N,u_N)$, which minimizes the loss function while satisfying the transition model as in Equation~(\ref{eqn:lqr_trans}). 

The optimal trajectory of the LQR problem is derived as,
\begin{align*}
u_t &= -F_tx_t\\
F_t &= (R_B^TP_tB)^{-1}(B^TP_tA),
\end{align*}
where $P_t$ is found by iteratively backwards in time,
\begin{align*}
P_{t-1} = A^TP_tA - (A^TP_tB)(R+B^TP_tB)^{-1}(B^TP_tA)+Q,
\end{align*}
with the terminal condition $P_n=Q$ \cite{skafsolving}.

\subsubsection{Iterative Linear Quadratic Regulator}
Iterative LQR (iLQR) is an iterative LQR method which finds the optimal trajectory by applying LQR repeatedly on the solved trajectory. In each iteration, iLQR adjusts the following cost function,
\begin{eqnarray*}
\lefteqn{l(x_{t+1}, u_{t+1}, \alpha_{t+1}) } \notag \\
&=& (1-\alpha)l(x_t,u_t) + \alpha(\lVert x_t - x^{(i)}_t  \rVert^2_2 + \lVert u_t - u^{(i)}_t\rVert^2_2) \notag
\end{eqnarray*}
where $l$ is a quadratic cost function. When $\alpha$ is close to one, the squared error term converges to zero and iLQR finds the optimal trajectory \cite{MPC_iLQR}.

By the trajectory optimization, it finds a trajectory $U^*$
which minimizes the sum of the cost function,
\begin{align*}
l_0(x,U) &= \sum\limits_{t=0}^{N-1} l(x_t,u_t) + l_f(x_N)\\
l_i(x,U) &= \sum\limits_{t=i}^{N-1} l(x_t,u_t) + l_f(x_N).
\end{align*}

\subsection{Deep Reinforcement Learning}
Deep reinforcement learning denotes reinforcement learning algorithms where the value functions are represented by deep neural networks \cite{Deep_RL}. Deep reinforcement learning models have achieved the state-of-the-art performance in various applications including a game playing console called, `Atari' \cite{AtariPaper}, by learning the complex control policies.

\subsubsection{Neural Networks for Reinforcement Learning}
The policy $\pi$ and the action-value function $Q^\pi(s,a)$ in the reinforcement learning is defined as follows.\\
\begin{displaymath}
a = \pi(s)
\end{displaymath}
\begin{displaymath}
Q^\pi(s,a) = \mathbb{E}[r_{t+1} + \gamma r_{t+2} + \gamma^2 r_{t+3} + ... |\ s,a],
\end{displaymath}
where $s$ and $a$ are respectively state and action, $r_t$ is the reward at t and $\gamma$ is the discount factor ranging from 0 to 1. Then, using the Bellman Equation and value iteration algorithms can solve above as, 
\begin{displaymath}
Q_{i+1}(s,a) = \mathbb{E}_{s'}[r + \gamma \max_{a'}Q_i(s',a') |\ s,a]
\end{displaymath}
Here, the action-value function Q can be represented by deep Q network with weight w,
\begin{displaymath}
Q_{w}(s,a) \approx Q^\pi(s,a).
\end{displaymath}

However, reinforcement learning using a nonlinear function approximator such as neural network  is known to be unstable or to diverge \cite{TD_approximation}. Therefore, recent variant of Q-learning suggests two key ideas: (1) playing a fixed target Q-network repeatedly; and (2) updating the target network parameters $w^-$ periodically  \cite{Deep_RL}. Using these two techniques, the Q-learning applying fixed parameters uses the following loss function
\begin{displaymath}
L(w) = \mathbb{E}_{(s,a,r,s') \sim U(D)}[(r + \gamma \max_{a'}Q(s',a',w^-)-Q(s,a,w))^2]
\end{displaymath}
where $D_{t} = \small\{ e_{1},...,e_{t} \small\} $ is a data set at each time-step $t$ storing plays of the agent, $e_{t} = (s_{t},a_{t},r_{t},s_{t+1})$ and $(s,a,r,s') \sim U(D)$ are the samples or minibatches of the plays.

\section{Learning Manipulations for Smart Devices}
\label{sec:learning_architecture}

Handling smart devices requires complex manipulations. Learning such controllers are non trivial due to dynamic changes of adequate pressures, velocities, and directions. In this work, we model the swipe actions of the smart device. 
  
\subsubsection{Our architecture of the manipulation}

Finding the optimal trajectory, from nonlinear dynamics, is not an easy task. Thus, we change the problem into a linear Gaussian model \cite{iLQR}. The swipe action control is locally trained with the linear dynamics by using iLQR. 

First, we initialize the five trajectory points for each direction; left, right, backward, and forward as plotted red in Fig. 3.

\begin{figure*}[t!]
\centering
\includegraphics[width=1\textwidth]{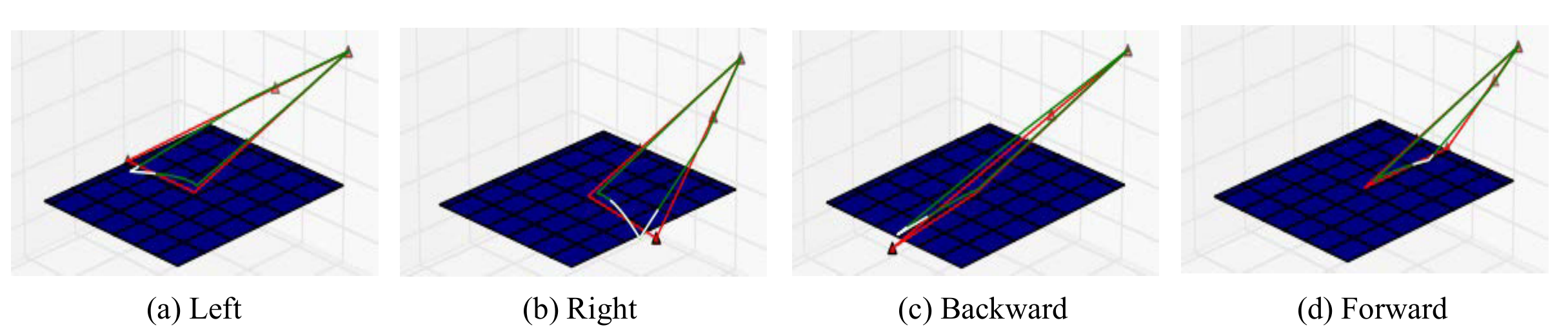}
\caption{Expected trajectory (red line) is just a linked line of input trajectory points (red triangle). After learning by iLQR, the optimized trajectory (green line) almost follows the expected trajectory (blue line). The white line is a set of points touching the surface (blue surface) which refers to the surface of the smart device with optimized trajectory.}
\end{figure*}

\subsubsection{Action and State Space for Baxter Research Robot}

Baxter research robot's action corresponds to the seven joint angles of its arm. In other words, the state $X_t$ is a set of the seven joint angles and the action $U_t$ is a set of the change of its joint angles. 
\begin{itemize}
\item State $X_t = 
[\theta^0_t,\cdots,\theta^6_t]$
\item Action $U_t = 
[\Delta \theta^0_t,\cdots,\Delta \theta^6_t]$
\end{itemize}

\subsubsection{Learning LQR for actual trajectories}
Considering the executable velocities of our Baxter robot, we discretize 3 seconds of time span into 250 steps. With more iterations conducted, the optimized trajectory by LQR fits to the intended trajectory.  
\begin{displaymath}
l_x = Q(X-X_N) +  \sum\limits_{t=1}^{N-1} Q(X-X_t),
\end{displaymath}
\begin{displaymath}
l_u = RU.
\end{displaymath}

\section{Learning to Play a Mobile Game for a Humanoid Robot}

Learning a smart phone game with the humanoid robot starts from learning its optimal policy in simulation at first. Since executing the game by the robot is relatively slower than executing it automatically, we have tried to find the optimal policy in simulation then, transfer it to the robot. 

We choose the game "\textit{2048}" \cite{the2048} because of its simplicity of playing and its complexity of winning the game. The input state is a bit string converted from a set of the positions of tiles and each value. The state is now with the 12 x 16 size since the number of possible values from 0 to 2048 (from $2^0$ to $2^{11}$) in 16 possible positions Fig. 2-(b.1). The output is one of four swiping actions: up, down, left, and right. The rewards are given to a sequence of actions when a tile marked with a goal number comes out.

As explained in Section~\ref{sec:experiments}-3, we have provided four heuristic algorithms for the "\textit{2048}" game to guide the action policy search. Note that, the heuristics are initial guides to help the reinforcement learning algorithm converge fast. The agent will not exactly follow the heuristics since the rewards from the heuristic algorithm are much lower than the ultimate reward for winning the game. The learning procedure consists of two steps: recognizing digits by neural networks and executing the actions provided by end-to-end architecture.

\subsubsection{Recognizing Digits by Neural Networks}
Since the inputs from the camera attached on the arm have noise by the direction or intensity of light, we did not give the input as the whole RGB image itself as done in Atari paper.  \cite{AtariPaper} Instead, we design a simple neural network to recognize digits in the smart phone since the digits recognition task in the smartphone resemble the digit recognition in the MNIST dataset \cite{MNIST}, an open set of hand-written digit images.

After the image segmentation in Fig. 2-(a.3), extracted tiles with the size of 32 by 32 pixels each enter into the neural network. To train the model with an expanded images, we gather the 14,034 tiles from the camera attached to the arm of the Baxter robot by randomly moving the robot for data augmentation. This data augmentation procedure makes a similar situation where the robot move as Fig. 2-(c) to (a). We also normalize data, and generated augmentation data by rotating and moving slightly \cite{DataAugmentation}. Now, the augmented data is with the 84,204 tiles. Such augmented data let our recognition accuracy as high as 98.9$\%$.

\begin{figure}[t]
\includegraphics[width=0.5\textwidth]{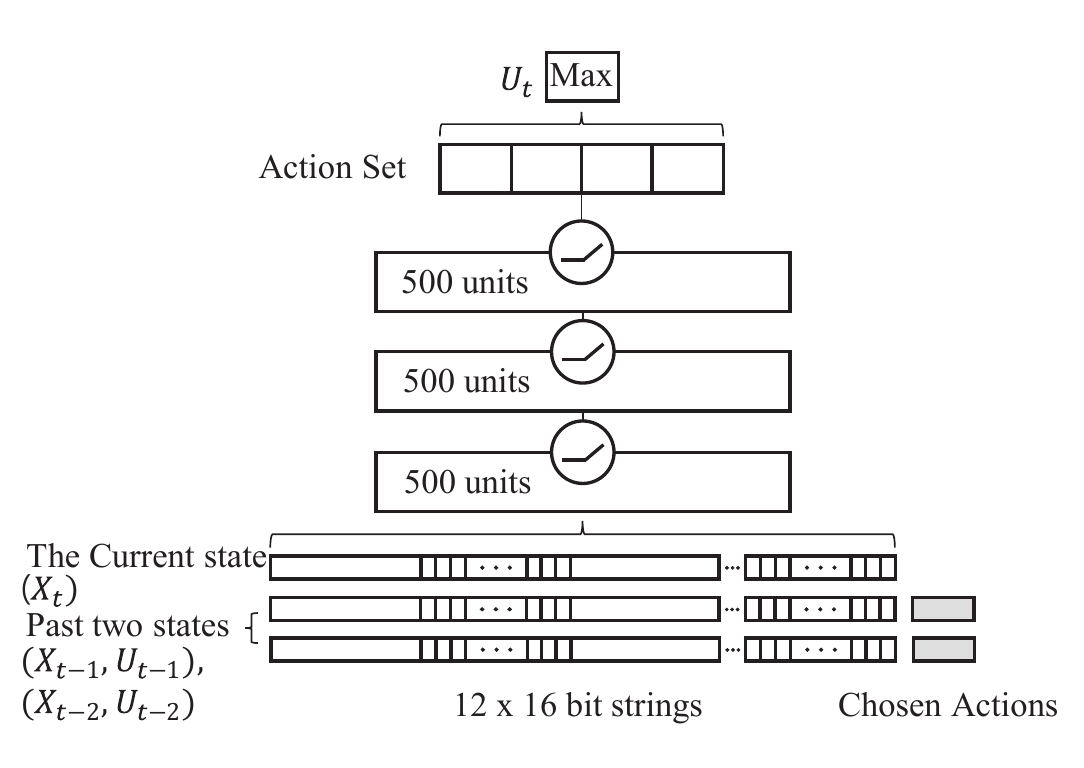}
\caption{A Deep-Q-Learning architecture diagram consisting of three layers. Each layer has 500 units. From the current state and past two (state, action) pairs, the network finds the optimal action.}
\end{figure}

\subsubsection{Executing the Actions by the Robot}
We design a deep Q-learning model as shown in Fig. 4. The DQL model approximates the Q function and learns to choose the best action out of four actions. As stated in the beginning of this section, it is hard to train the network for solving the game by taking actions from the robot itself; every action by the robot takes more than a second, which is quite slow. Therefore, we first train the deep neural network in the simulator where actions takes less than 0.01 seconds, and then apply the trained network into the robot. Since input digits, which are recognized by the neural network are discrete numbers from 0 to 11, we changed the set into a bit string which indicates the specific value in a specific position as 1 or 0. The bitmap is size of 16 x 12, in ascending order from 0 to 2048 for 16 positions, and the same value of tiles are adjacently mapped as shown in Fig. 2-(b.1). We use three layers of fully connected neural network, each of them consists of 500 hidden units and the action set includes four swipe actions: Up, Down, Left, and Right. Finally, the learning step is composed of three steps: forward, action, and backward as follows.
 
\textbf{Forward}: Every previous two pairs (state and action) is saved in our memory to fill up the input for our network. Based on the previous two states, action pairs and the current state, the reward is calculated for each action. An action with the maximum reward is selected as the best action. We use $\varepsilon$-greedy algorithm \cite{epsilon_greedy} where $\varepsilon$ which determines the probability of choosing an random action (instead of best action chosen from the network), decreases from 1.0 to 0.05 as age increases in each 2,000 learning iteration.

\textbf{Action}: After choosing the best action from the policy network, we get the output from the game. Especially, in our \textit{"2048"} game, the computer (game) inserts randomly a 2 or 4 tile in an empty space (adversarial moves).

\textbf{Backward}: We give the reward computed from the last action and a set of states ($s_t$, $a_t$, $s_{t+1}$). In addition, the network will be trained by random sampling which selects samples from our entire plays. The batch size is 16 and the total playing size is 30,000. In each backward step, we append the current state, action, reward and the next state, ($s_t$, $u_t$, $r_t$, $s_{t+1}$). When the playing history is full, one of the existing entry is replaced by a new entry.

\section{Experimental Results}
\label{sec:experiments}


\subsubsection{Details of the "2048" game}
The \textit{"2048"} game \cite{the2048} is an variant of a preceding game of the Threes \cite{threes}. Given a 4 by 4 grid, the player can move the grid in 4 directions (up, down, right and left). Each action pushes all tiles to one direction and makes the adjacent tiles to be merged when they have the same value as shown Fig. 5. The merged tile now has the sum of two values. After every action, computer places 2 or 4 in each empty tile with the probability of 0.9 and 0.1, respectively.

\begin{figure}[t]
\centering
\label{2048_merge}
\includegraphics[width=0.4\textwidth]{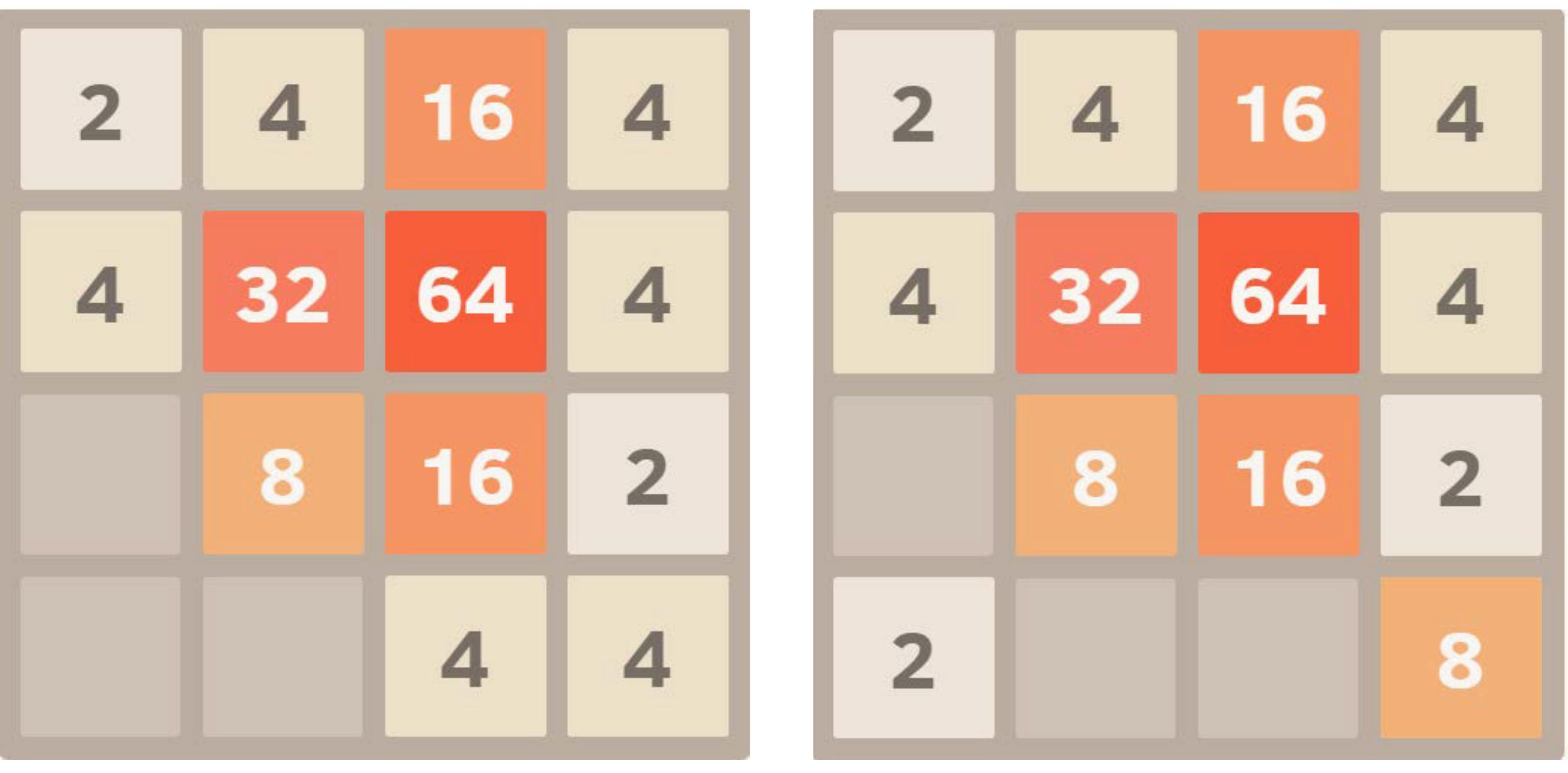}
\caption{The \textit{"2048"} game board is with a 4 by 4 grid. The image in the right-hand side is generated by taking the right (move) action from the image in the left-hand side. Two values of 4 tiles in the last row are merged into 8. A tile with a value 2 is generated at the left-bottom corner.}
\end{figure}

\subsubsection{Heuristics of the "2048" game}
The four typical heuristics in the \textit{"2048"} game are presented as follows.

\textbf{Monotonicity}: This heuristic measures whether the values of the tiles are strictly increasing or decreasing along the pair of the horizontal directions (left/right) and the vertical directions directions (up/down). It tries to make the board well organized where larger numbers are located in the corners by merging tiles with smaller numbers on the different sides.

\textbf{Smoothness}: It is from an idea that the adjacent tiles with the same number are merged. It calculates the value difference between its neighbors in four directions and tries to minimize the difference. 

\textbf{Free tiles}: It gives a penalty when there are only a few free tiles left since we lose the game when there is no free space. 

\textbf{Maximum value}: It calculates the maximum value among the whole tiles.

\subsubsection{Results and Discussions}

As shown in Table~\ref{tbl:result}, our Baxter robot reaches the number 128 with winning rate more than 90 percent which is comparable to the experiment in a simulated environment. The result is significantly better than the performance of random moves strategy which is only 53.98$\%$. The wining rate of the game for achieving the number 256 is now 54.00$\%$ which is comparable to the Deep Q learning model in a simulated environment. Note that the performance of random moves is just 7.09$\%$. 

During the robotic experiments, there are two factors which may incur errors beyond learning game policy model. First, our recognition model (neural network) successfully recognized digits with only 1.1$\%$ error. However, the recognition errors in the robotic experiments could be higher. The iLQR controller trained in the simulated environment could have erroneous actions in the Baxter experiments. Table~\ref{tbl:result} presents reflects the errors (difference) rate incurred from two potential erroneous factors. The difference of simulation and Baxter executions with respect to the number of iterations per game is about $7\%$. Table~\ref{tbl:result_iter} reflects the accumulated erroneous moves caused by the recognition errors and control errors. 
\begin{table}[t!]
\centering
\caption{Comparisons of Winning Rate of the \textit{"2048"} game}
\label{tbl:result}
\begin{tabular}{|c||r|r|r|r|}
\hline
Target Value& \thead{Random\\Moves} &  \thead{Deep Q\\(Simulator)} & \thead{Deep Q\\(Baxter)} & \thead{difference} \\ 
\hline
128 & 53.98\% & 92.81\% &90.74\% & -2.07\% \\
\hline
256 & 7.09\% & 54.35\% &54.00\%  &-0.35\% \\       
\hline
\end{tabular}
\end{table}

\begin{table}[t!]
\centering
\caption{Comparisons of the number of iterations per game}
\label{tbl:result_iter}
\begin{tabular}{|c||r|r|r|r|}
\hline
Target Value&  \thead{Deep Q\\(Simulator)} & \thead{Deep Q\\(Baxter)} & difference(\%) \\ 
\hline
128  & 86 & 90 & 4.65\% \\
\hline
256  & 146 & 157 & 7.53\%  \\       
\hline
\end{tabular}
\end{table}

\begin{figure*}[t!]
\centering
\label{experiment}
\includegraphics[width=1\textwidth]{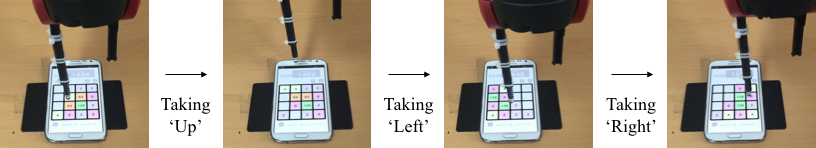}
\caption{A series of playing \textit{"2048"} game by the robot Baxter with the trained model. The leftmost figure is the one after taking 150th actions. Then, by taking 'UP' action  from the leftmost one, there are two '64' tiles (orange) lined up horizontally.
By taking 'Left' action, they are merged to a '128' tile (green) as shown in the 3rd figure. Finally, there come  two '128' tiles standing in a line vertically by taking 'Right' action.}
\end{figure*}

\section{CONCLUSION AND FUTURE WORK}
\label{sec:conclusion}

Deep learning may boost the performance of robotic manipulation  when non-physical state changes of objects are important. In this paper, we present an architecture diagram which seamlessly combines the recognition, planning and execution. We demonstrate how deep learning can improve the performance of recognition and planning in a humanoid robot. In the experiments with the Baxter Research robot, we show that the Baxter research robot equipped with our architecture diagram could achieve the high winning rate of a complex mobile game, \textit{"2048"} game, as the Deep Q learning algorithm in simulation. 

\section*{ACKNOWLEDGMENT}
This work was supported by Basic Science
Research Program through the National Research Foundation of Korea (NRF) grant funded by the Ministry of Science, ICT \& Future Planning (MSIP) (NRF- 2014R1A1A1002662), the NRF grant funded by the MSIP (NRF-2014M2A8A2074096) and supported by the Industrial Convergence Core Technology Development Program(No. 10063172) funded by MOTIE, Korea. Authors thank Phuong Hoang and Janghoon Ju for their constructive comments.

\bibliographystyle{IEEEtran}
\bibliography{references}

\begin{thebibliography}{10}
\providecommand{\url}[1]{#1}
\csname url@samestyle\endcsname
\providecommand{\newblock}{\relax}
\providecommand{\bibinfo}[2]{#2}
\providecommand{\BIBentrySTDinterwordspacing}{\spaceskip=0pt\relax}
\providecommand{\BIBentryALTinterwordstretchfactor}{4}
\providecommand{\BIBentryALTinterwordspacing}{\spaceskip=\fontdimen2\font plus
\BIBentryALTinterwordstretchfactor\fontdimen3\font minus
  \fontdimen4\font\relax}
\providecommand{\BIBforeignlanguage}[2]{{%
\expandafter\ifx\csname l@#1\endcsname\relax
\typeout{** WARNING: IEEEtran.bst: No hyphenation pattern has been}%
\typeout{** loaded for the language `#1'. Using the pattern for}%
\typeout{** the default language instead.}%
\else
\language=\csname l@#1\endcsname
\fi
#2}}
\providecommand{\BIBdecl}{\relax}
\BIBdecl

\bibitem{Control}
S.~Levine, C.~Finn, T.~Darrell, and P.~Abbeel, ``End-to-end training of deep
  visuomotor policies,'' \emph{CoRR}, vol. abs/1504.00702, 2015.

\bibitem{toy}
S.~Levine, N.~Wagener, and P.~Abbeel, ``Learning contact-rich manipulation
  skills with guided policy search,'' in \emph{{IEEE} International Conference
  on Robotics and Automation, {ICRA} 2015, Seattle, WA, USA, 26-30 May, 2015},
  2015, pp. 156--163.

\bibitem{AtariPaper}
V.~Mnih, K.~Kavukcuoglu, D.~Silver, A.~Graves, I.~Antonoglou, D.~Wierstra, and
  M.~A. Riedmiller, ``Playing atari with deep reinforcement learning,''
  \emph{CoRR}, vol. abs/1312.5602, 2013.

\bibitem{LQRIROS15}
W.~Han, S.~Levine, and P.~Abbeel, ``Learning compound multi-step controllers
  under unknown dynamics,'' in \emph{2015 {IEEE/RSJ} International Conference
  on Intelligent Robots and Systems (IROS)}, 2015, pp. 6435--6442.

\bibitem{baxter}
E.~Guizzo and E.~Ackerman, ``The rise of the robot worker,'' \emph{IEEE
  Spectrum}, vol.~49, no.~10, pp. 34--41, 2012.

\bibitem{the2048}
\BIBentryALTinterwordspacing
S.~Rodriguez, ``Maker of hit puzzle game '2048' says he created it over a
  weekend,'' \emph{The Los Angeles Times}, March 11 2014. [Online]. Available:
  \url{http://www.latimes.com/}
\BIBentrySTDinterwordspacing

\bibitem{adaptivegaussian}
M.~Teutsch, P.~Trantelle, and J.~Beyerer, ``Adaptive real-time image smoothing
  using local binary patterns and gaussian filters,'' in \emph{{IEEE}
  International Conference on Image Processing, {ICIP} 2013, Melbourne,
  Australia, September 15-18, 2013}, 2013, pp. 1120--1124.

\bibitem{intel_robot}
T.~Simonite, ``Intel robot puts touch screens through their paces,'' \emph{MIT
  Technology Review}.

\bibitem{tapster}
\BIBentryALTinterwordspacing
K.~Finley, ``Robot with long finger wants to touch your iphone apps,'' August 1
  2013. [Online]. Available: \url{http://www.wired.com/2013/08/tapster/}
\BIBentrySTDinterwordspacing

\bibitem{chameleon}
\BIBentryALTinterwordspacing
P.~Electronics. (2015) Agile testing for electronic devices. [Online].
  Available: \url{http://www.pkcelectronics.com/}
\BIBentrySTDinterwordspacing

\bibitem{optofidelity}
\BIBentryALTinterwordspacing
OptoFidelity. (2015) Phone validation by optofidelity human simulator robot.
  [Online]. Available: \url{http://www.optofidelity.com}
\BIBentrySTDinterwordspacing

\bibitem{LQR_ex1}
D.~H. Nguyen, ``A sub-optimal consensus design for multi-agent systems based on
  hierarchical {LQR},'' \emph{Automatica}, vol.~55, pp. 88--94, 2015.

\bibitem{LQR_ex2}
A.~Mosebach and J.~Lunze, ``{LQR} design of synchronizing controllers for
  multi-agent systems,'' \emph{Automatisierungstechnik}, vol.~63, no.~6, pp.
  403--412, 2015.

\bibitem{LQR_ex3}
A.~M. Benomair, F.~A. Bashir, and M.~O. Tokhi, ``Optimal control based
  lqr-feedback linearisation for magnetic levitation using improved spiral
  dynamic algorithm,'' 2015, pp. 558--562.

\bibitem{iLQR}
H.~Zhang, J.~Gong, Y.~Jiang, G.~Xiong, and H.~Chen, ``An iterative linear
  quadratic regulator based trajectory tracking controller for wheeled mobile
  robot,'' \emph{Journal of Zhejiang University - Science {C}}, vol.~13, no.~8,
  pp. 593--600, 2012.

\bibitem{MPC_iLQR}
Y.~Tassa, T.~Erez, and E.~Todorov, ``Synthesis and stabilization of complex
  behaviors through online trajectory optimization,'' in \emph{2012 {IEEE/RSJ}
  International Conference on Intelligent Robots and Systems (IROS)}, 2012, pp.
  4906--4913.

\bibitem{Tennis10}
J.~Kober, E.~Oztop, and J.~Peters, ``Reinforcement learning to adjust robot
  movements to new situations,'' in \emph{Robotics: Science and Systems VI
  (RSS)}, 2010.

\bibitem{Deisenroth11}
M.~P. Deisenroth, C.~E. Rasmussen, and D.~Fox, ``Learning to control a low-cost
  manipulator using data-efficient reinforcement learning,'' in \emph{Robotics:
  Science and Systems VII (RSS)}, 2011.

\bibitem{KaelblingL13}
L.~P. Kaelbling and T.~Lozano{-}P{\'{e}}rez, ``Integrated task and motion
  planning in belief space,'' \emph{I. J. Robotic Res.}, vol.~32, no. 9-10, pp.
  1194--1227, 2013.

\bibitem{KonidarisKL14}
G.~Konidaris, L.~P. Kaelbling, and T.~Lozano{-}P{\'{e}}rez, ``Constructing
  symbolic representations for high-level planning,'' in \emph{Proceedings of
  the Twenty-Eighth {AAAI} Conference on Artificial Intelligence (AAAI)}, 2014,
  pp. 1932--1938.

\bibitem{taskmotion}
L.~P. Kaelbling and T.~Lozano-Pérez, ``Hierarchical task and motion planning
  in the now,'' in \emph{2011 IEEE International Conference on Robotics and
  Automation (ICRA)}, 2011, pp. 1470--1477.

\bibitem{CPMP09}
J.~Choi and E.~Amir, ``Combining planning and motion planning,'' in \emph{2009
  {IEEE} International Conference on Robotics and Automation (ICRA)}, 2009, pp.
  238--244.

\bibitem{skafsolving}
J.~Skaf, ``Solving the lqr problem by block elimination.''

\bibitem{Deep_RL}
V.~Mnih, K.~Kavukcuoglu, D.~Silver, A.~A. Rusu, J.~Veness, M.~G. Bellemare,
  A.~Graves, M.~Riedmiller, A.~K. Fidjeland, G.~Ostrovski, S.~Petersen,
  C.~Beattie, A.~Sadik, I.~Antonoglou, H.~King, D.~Kumaran, D.~Wierstra,
  S.~Legg, and D.~Hassabis, ``Human-level control through deep reinforcement
  learning,'' \emph{Nature}, 2015.

\bibitem{TD_approximation}
\BIBentryALTinterwordspacing
V.~Tadic, ``Convergence analysis of temporal-difference learning algorithms
  with linear function approximation,'' in \emph{Proceedings of the Twelfth
  Annual Conference on Computational Learning Theory, {COLT} 1999, Santa Cruz,
  CA, USA, July 7-9, 1999}, 1999, pp. 193--202. [Online]. Available:
  \url{http://doi.acm.org/10.1145/307400.307438}
\BIBentrySTDinterwordspacing

\bibitem{MNIST}
L.~Deng, ``The {MNIST} database of handwritten digit images for machine
  learning research [best of the web],'' \emph{{IEEE} Signal Process. Mag.},
  vol.~29, no.~6, pp. 141--142, 2012.

\bibitem{DataAugmentation}
Z.~Gan, R.~Henao, D.~E. Carlson, and L.~Carin, ``Learning deep sigmoid belief
  networks with data augmentation,'' in \emph{Proceedings of the Eighteenth
  International Conference on Artificial Intelligence and Statistics, {AISTATS}
  2015, San Diego, California, USA, May 9-12, 2015}, 2015.

\bibitem{epsilon_greedy}
C.~J. C.~H. Watkins, ``Learning from delayed rewards,'' {PhD} dissertation,
  University of Cambridge, Cambridge, England, 1989.

\bibitem{threes}
\BIBentryALTinterwordspacing
B.~Kuchera. (2015) Why it took a year to make, and then break down, an amazing
  puzzle game. [Online; posted 06-February-2013]. [Online]. Available:
  \url{http://www.polygon.com/2014/2/6/5386200/why-it-took-a-year-to-make-and-then-break-down-an-amazing-puzzle-game}
\BIBentrySTDinterwordspacing

\end{thebibliography}
\end{document}